\documentclass[letterpaper, 10 pt, conference]{ieeeconf}  

\IEEEoverridecommandlockouts                              

\newlength\figureheight             		
\newlength\figurewidth 						

\usepackage{graphicx}
\graphicspath{{figures/}}
\usepackage{tikz}							
\usepackage{pgfplots}						
\pgfplotsset{compat=newest}					
\usepackage{siunitx}						
\usepackage{booktabs}						
\usepackage{mathtools} 						
\usepackage{nicefrac} 						
\usepackage{multirow} 						
\usepackage{subcaption}						
\usepackage{cite}							

\newcommand{\ego}{ego-vehicle }				
\newcommand{\tabelle}{Table~}				
\newcommand{\abbildung}{Fig.~}				
\newcommand{\gleichung}{Equation~}			
\newcommand{\gleichungen}{Equations~}		
\newcommand{\radar}{Radar }					
\newcommand{\radarohne}{Radar}				
\newcommand{\lidar}{Lidar }					
\newcommand{\lidarohne}{Lidar}				
\newcommand{\snr}{signal-to-noise ratio }	
\newcommand{\snro}{signal-to-noise ratio}	

\usetikzlibrary{external}
\usepackage{textcomp}
\usepackage{lipsum}
\newcommand\copyrighttext{%
	\footnotesize \textcopyright 2019 IEEE.  Personal use of this material is permitted.  Permission from IEEE must be obtained for all other uses, in any current or future media, including reprinting/republishing this material for advertising or promotional purposes, creating new collective works, for resale or redistribution to servers or lists, or reuse of any copyrighted component of this work in other works.
}
\newcommand\copyrightnotice{%
	\tikzset{external/export=false}
	\begin{tikzpicture}[remember picture,overlay]
	\node[anchor=south,yshift=10pt, xshift=10pt] at (current page.south) {\fbox{\parbox
			{\dimexpr\textwidth-\fboxsep-\fboxrule\relax}{\copyrighttext}}};
	\end{tikzpicture}%
	\tikzset{external/export=true}
}

\title{\LARGE \bf
	Systematic Analysis of the Sensor Coverage of Automated Vehicles Using Phenomenological Sensor Models*
}

\author{Thomas Ponn$^{1}$, Fabian M\"uller$^{1}$ and Frank Diermeyer$^{1}$
\thanks{*The research project was funded and supported by T\"UV S\"UD Autoservice GmbH.}
\thanks{$^{1}$Thomas Ponn, Fabian M\"uller and Frank Diermeyer are with the Institute of Automotive Technology, Technical University of Munich, 85748 Garching, Germany
        {\tt\small \{ponn,diermeyer\}@ftm.mw.tum.de}, \tt\small fabian93.mueller@tum.de}%
}

\begin{document}

\maketitle
\thispagestyle{empty}
\pagestyle{empty}
\copyrightnotice

\begin{abstract}

The objective of this paper is to propose a systematic analysis of the sensor coverage of automated vehicles. Due to an unlimited number of possible traffic situations, a selection of scenarios to be tested must be applied in the safety assessment of automated vehicles. This paper describes how phenomenological sensor models can be used to identify system-specific relevant scenarios. In automated driving, the following sensors are predominantly used: camera, ultrasonic, \radar and \lidarohne.  Based on the literature, phenomenological models have been developed for the four sensor types, which take into account phenomena such as environmental influences, sensor properties and the type of object to be detected. These phenomenological models have a significantly higher reliability than simple ideal sensor models and require lower computing costs than realistic physical sensor models, which represents an optimal compromise for systematic investigations of sensor coverage. The simulations showed significant differences between different system configurations and thus support the system-specific selection of relevant scenarios for the safety assessment of automated vehicles. 

\end{abstract}


\section{INTRODUCTION}
\label{sec:introduction}

In addition to the development of automated driving functions, proving that they are safe is one of the most relevant tasks facing the entire automotive industry. Proof of safety is of particular importance for systems of automation level 3 and higher according to SAE \cite{SAEJ3016.2016} and remains an unsolved problem. The difference between lower levels of automation and level 3 (or higher) is that the responsibility for the driving task transfers from the human driver to the system. In contrast to level 2, this means that the driver no longer has to permanently monitor the system. At level 2, the driver is required to intervene immediately if the driving function responds incorrectly and the driver has to establish a safe driving condition. For level 3 and higher, this means that the activated system must be able to handle all traffic situations independently and safely because the driver can no longer be used as a fall-back option. Due to the open-parameter space that occurs in reality, an infinite number of scenarios can theoretically be defined. An economic proof of the safety of automated vehicles is therefore not possible.

In order to achieve an economic safety assessment of automated driving functions, a method for determining relevant scenarios must be developed. One approach for this is so-called scenario-based testing, which is also being developed in the German funding project PEGASUS \cite{DeutschesZentrumfurLuftundRaumfahrte.V..2018}. Based on the assumption that the majority of situations occurring in real traffic are uncritical, scenario-based testing focuses on relevant, critical scenarios, thus reducing the test scope. A reduction of costs is achieved by increasing the use of simulation. The previously unsolved task of identifying all relevant scenarios remains, despite this approach. 

When selecting the relevant scenarios to be tested, many publications focus on testing the driving function using ground-truth data \cite{Althoff.2009, Gruber.2018, Madrigal.2017}. The perception of the environment by sensors is neglected, but this has a significant influence on the performance of the overall system. A scenario based on ground-truth data can be uncritical for the driving function, but can lead to risks or accidents if sensor uncertainties are taken into account (e.g. Uber accident \cite{NTSB.2018}). Since each manufacturer uses an individual sensor setup, it follows that the selection of the relevant scenarios must also be adapted to the respective system under test. 

This contribution therefore presents a novel approach for a systematic analysis of the sensor coverage of automated vehicles, which can be used to derive relevant test cases for the system to be tested. The results show that different system configurations have clearly different weaknesses, which have to be considered during the tests.

The article is structured as follows: Section \ref{subsec:sensor_models} introduces different types of sensor models. Based on this, Section \ref{subsec:sensor_fundamentals} explains the basics of phenomenological sensor models. Section \ref{sec:methodology} describes in detail the procedure for modeling the sensor coverage. In the results (Section \ref{sec:results}), two system configurations are compared and system-specific relevant test scenarios are derived. The results are then critically discussed in Section \ref{sec:discussion}. A summary (Section \ref{sec:conclusion}) including an outlook for future work concludes the article.

\section{RELATED WORK}
\label{sec:related_work}

The following section gives an overview of sensor models used in literature and explains the basics of the sensors used in the automotive industry. 

\subsection{Sensor Models}
\label{subsec:sensor_models}

Sensor models can be used for simulation-based testing of automated vehicles and for sensor coverage analysis. In literature, there are different types of sensor models, which differ in their level of detail. The choice of the optimal level of detail depends on the intended use and must therefore be adjusted accordingly. In the following, three sensor models, sorted according to the increasing level of detail, are explained: ideal, phenomenological and physical sensor models. The classification and the explanations are based on \cite{Bernsteiner.2015, Feilhauer.2017b, Schubert.2014, Cao.2017}.

\subsubsection{Ideal Sensor Model}
\label{subsubsec:ideal_sensor_model}

Ideal sensor models map the geometric space of the sensor coverage without measurement errors, i.e. objects are detected any time when they are within the sensor's field of view. Physical effects are not taken into account. Idealized sensor models represent so-called ground-truth models that provide the true, undisturbed values of the simulated quantities (e.g. position of an object).

\subsubsection{Phenomenological Sensor Model}
\label{subsubsec:phenomenological_sensor_model}

Phenomenological sensor models simulate the properties of the sensors and real effects. Physical effects are modeled phenomenologically, i.e. the result of the effects is reproduced. The relationship between inputs and outputs of the sensors as well as the exact internal processes and effects are unknown.

\subsubsection{Physical Sensor Model}
\label{subsubsec:physical_sensor_model}

Physical sensor models map the characteristics of the sensors, the physical principle and physical effects correctly. One physical modeling approach is the so-called ray-tracing model that uses the ray-tracing approach. It is assumed that the transmitted and received signals propagate along rays which are reflected and refracted on objects based on the laws of reflection and refraction.

The characteristics of the different sensor models are summarized in \tabelle \ref{tab:sensor_models}. Phenomenological sensor models represent the optimal trade-off between level of detail and computing costs for analyzing the sensor coverage of automated vehicles. 

\begin{table}[b]
	\caption{Characteristics of different sensor models.}
	\label{tab:sensor_models}
	\begin{center}
		\begin{tabular}{rcc}
			\cmidrule[0.1em](){2-3} 
			& Level of detail 	& Computing costs\\
			\midrule
			Ideal 				& low 				& low \\
			Phenomenological 	& mid 				& mid \\
			Physical 			& high 				& high \\
			\bottomrule
		\end{tabular}
	\end{center}
\end{table}

\subsection{Sensor Fundamentals}
\label{subsec:sensor_fundamentals}

\begin{figure}[t]
	\centering
	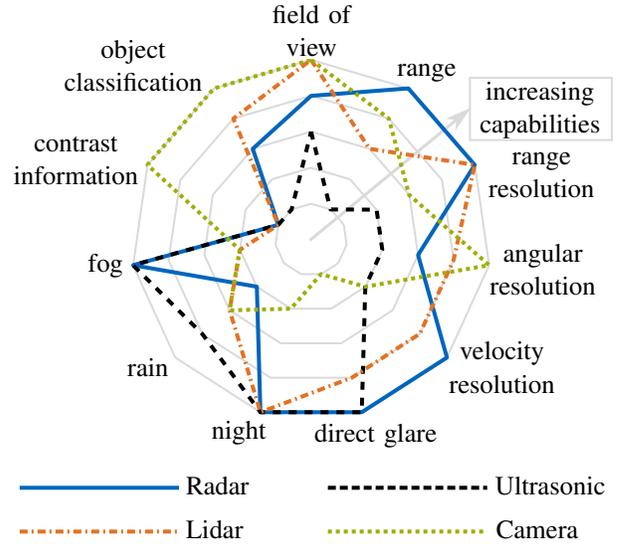
	\caption{Strength and weaknesses of the most important sensors for automated driving (adapted from \cite{Steinbaeck.2017} and \cite{Schrepfer.2018}).}
	\label{fig:sensor_comparison}
\end{figure}

The most important sensor types for automated driving are ultrasonic, \radarohne, \lidar and camera \cite{Steinbaeck.2017}. By combining these different types of sensors, sensor-specific weaknesses of individual types can be compensated (\abbildung \ref{fig:sensor_comparison}).

In the following, the basics for the description of phenomenological models of these four sensor types are explained. Based on their functional principles, the sensor types can be divided into active and passive sensors. While active sensors emit a signal and receive the signal reflected by the object, passive sensors do not emit a signal and only measure signals emitted by the object. Ultrasonic, \radar and \lidar can be classified as active sensors, and cameras as passive sensors. The symbols used in the following Equations and their descriptions can be seen in \tabelle \ref{tab:sensor_symbols}.

\begin{table}[!b]
	\caption{List of symbols, their unit and description.}
	\label{tab:sensor_symbols}
	\begin{center}
		\begin{tabular}{ccl}
			\toprule 
			Symbol & Unit 	& Description \\
			\midrule
			$A_{\text{r},i} $		& \SI{}{\square\meter} 				& receiver area of sensor $i$ \\
			$B_{\text{n},i} $		& \SI{}{\nicefrac{1}{\second}} 		& noise bandwidth of sensor $i$ \\
			$D_{\text{r}} $			& \SI{}{\meter} 					& receiver diameter \\
			$E_{\text{sun}} $		& \SI{}{\nicefrac{\watt}{\meter^2}}	& radiation density of sunlight \\
			$G_{\text{e},i} $		& - 								& emitter antenna gain factor sensor $i$ \\
			$G_{\text{r},i} $		& - 								& receiver antenna gain factor sensor $i$ \\
			$h $					& \SI{}{\joule \second} 			& Planck constant \\
			$i $					& - 								& type of sensor: ultrasonic, \radarohne, \lidar \\
			$k $					& \SI{}{\nicefrac{\joule}{\kelvin}} & Boltzmann constant \\
			$L_{\text{oa},i} $		& -									& overall damping coefficient of sensor $i$ \\
			$N_{\text{e}}$ 			& - 								& number of generated electrons\\
			$N_{\text{e,sun}}$ 		& - 								& number of generated electrons by direct glare\\
			$N_{\text{pix,camera}}$ & - 								& number of pixel of the camera\\
			$N_{\text{pix,o}} $		& - 								& object size in the image measured in pixel \\
			$P_{\text{e},i} $		& \SI{}{\watt} 						& sensor $i$ emitting power \\
			$P_{\text{n},i} $		& \SI{}{\watt} 						& sensor $i$ noise power \\
			$P_{\text{r},i} $		& \SI{}{\watt} 						& sensor $i$ receiving power \\
			$R $					& \SI{}{\meter} 					& distance \\
			$QE $					& -									& quantum efficiency at photon conversion \\
			$SNR_{i} $				& - 								& \snr of sensor $i$ \\
			$t_{\text{int}} $		& \SI{}{\second} 					& integration time of camera sensor \\
			$T_{\text{sys},i} $		& \SI{}{\kelvin} 					& system noise temperature of sensor $i$ \\
			$\Theta_{\text{lidar}} $& \SI{}{\radian} 					& laser-beam expanding angle of \lidar sensor \\
			$\lambda_{i} $			& \SI{}{\meter} 					& wavelength of emitted signal of sensor $i$ \\
			$\nu $					& \SI{}{\nicefrac{1}{\second}} 		& radiation frequency \\
			$\sigma_{i} $			& \SI{}{\meter^2} 					& object cross-section regarding sensor $i$ \\
			$\phi $					& \SI{}{\radian} 					& azimuth angle \\
			$\psi $					& \SI{}{\radian} 					& elevation angle \\
			\bottomrule
		\end{tabular}
	\end{center}
\end{table}

\subsubsection{Active Sensors}
\label{subsubsec:active_sensors}

The active sensors - ultrasonic, \radar and \lidar - operate according to a similar principle: high-frequency radiation (sound, radio waves or laser-beams) is emitted by the transmitter and reflected by objects. The reflected radiation is absorbed by the receiver and processed further. Signal processing can be used to obtain information about the detected object (e.g. the distance).

In general, the performance of sensors can be expressed by the \snr $SNR$ that describes the ratio between the power of the received signal $P_{\text{r}}$ and the power of the noise $P_{\text{n}}$. It can be calculated according to \gleichung \ref{eq:SNR_general}.

\begin{equation}
SNR = \frac{P_{\text{r}} }{P_{\text{n}} }
\label{eq:SNR_general}
\end{equation}

The received power $P_{\text{r}}$ for ultrasonic and \radar sensors can be calculated identically (\gleichung \ref{eq:P_reflected}). A detailed derivation of this formula can be found in \cite{Sonbul.2014} for ultrasonic and in \cite{Grgic.2015, Ludloff.1998, Blake.1986} for \radar sensors.

\begin{align}
P_{\text{r},i}(R,\phi,\psi) &= \frac{P_{\text{e},i} G_{\text{e},i}(\phi,\psi) G_{\text{r},i}(\phi,\psi) \sigma_i \lambda_i^2}{(4\pi)^3 R^4 L_{\text{oa},i}(R)}  \notag \\
&\text{with} \qquad i=\{\text{ultrasonic; \radarohne}\}
\label{eq:P_reflected}
\end{align}

The noise power $P_{\text{n},i}$ can also be expressed identically for ultrasonic and \radar sensors and depends on the Boltzmann constant $k$, the noise bandwidth $B_{\text{n},i}$ of sensor $i$ and the system noise temperature $T_{\text{sys},i}$ of sensor $i$. According to \cite[p. 42]{Ludloff.1998}, the noise bandwidth $B_{\text{n},i}$ can be approximated by $\tau_i^{-1}$, where $\tau_i$ corresponds to the pulse width of the transmitted signal of sensor $i$. The system noise temperature $T_{\text{sys},i}$ represents all internal system losses of sensor $i$ \cite[p. 17]{Ludloff.1998}. In combination with \gleichung \ref{eq:SNR_general} and \gleichung \ref{eq:P_reflected} the \snr of ultrasonic and \radar sensor can be calculated according to \gleichung \ref{eq:SNR_radar}.

\begin{equation}
SNR_i(R, \phi,\psi) = \frac{P_{\text{r},i} (R, \phi,\psi) }{k B_{\text{n},i} T_{\text{sys},i}} 
\label{eq:SNR_radar}
\end{equation}

For \lidar sensors, \gleichung \ref{eq:SNR_general} can again be used as a starting point. The calculation of the received power $P_{\text{r,lidar}}$ as well as the noise power $P_{\text{n,lidar}}$ differs from the calculation of ultrasonic and \radar sensors. The following equations are based on \cite{Richmond.2010, TristanK..2010, Sameh.2016, Hayat.2000}, which use an adapted \radar equation for determining the received power $P_{\text{r,lidar}}$. Using small-angle approximation for laser-beam expanding angle $\Theta_{\text{lidar}}$, the received power $P_{\text{r,lidar}}$ can be calculated according to \gleichung \ref{eq:P_reflected_lidar}.

\begin{equation}
P_{\text{r,lidar}}(R) = \frac{P_{\text{e,lidar}} \sigma_{\text{lidar}} A_{\text{r,lidar}}}{\pi^2 R^4 \Theta_{\text{lidar}}^2 L_{\text{oa,lidar}}(R)}
\label{eq:P_reflected_lidar}
\end{equation}

The noise power $P_{\text{n,lidar}}$ for \lidar sensors in the automotive sector is primarily dominated by shot noise \cite[p. 3]{TristanK..2010}, \cite[p. 2]{Sameh.2016}. Thermal noise and background noise recorded in the form of temperature radiation can be represented by the system noise temperature $T_{\text{sys,lidar}}$ and takes internal system losses into account. The first summand in \gleichung \ref{eq:P_noise_lidar} corresponds to shot and the second one to thermal and background noise, respectively.

\begin{equation}
P_{\text{n,lidar}} = 2 h \nu B_{\text{n,lidar}} + k B_{\text{n,lidar}} T_{\text{sys,lidar}}
\label{eq:P_noise_lidar}
\end{equation}

\subsubsection{Passive Sensors}
\label{subsubsec:passive_sensors}

Cameras as passive sensors work according to a different functional principle: instead of actively emitting radiation, a receiver absorbs secondary high-frequency radiation in the visible range (light). Then, as with active sensors, information about the environment can be obtained with the aid of signal processing \cite[p. 6]{Rudolph.2017}. If the sun or light in general is regarded as the transmitter of camera sensors, the operating principle can be approximated to that of active sensors.

Due to the different working principle of the camera, the \snr cannot be expressed by the ratio of the received radiation to the noise signal. To get a comparable approach for the camera, its \snr is expressed by the ratio of electrons generated by the object in the vehicle's environment to the electrons generated by the illumination of the remaining environment. The following equations are based on \cite[chap. 3f.]{Lange.2000}. First, the power received by the camera $P_{\text{r,camera}}$ is calculated (\gleichung \ref{eq:P_reflected_camera}). 

\begin{equation}
P_{\text{r,camera}}(R) = \frac{E_{\text{sun}} \sigma_{\text{camera}} D_{\text{r}}^2 N_{\text{pix,o}}(R)}{16 R^2 L_{\text{oa,camera}} (R) N_{\text{pix,camera}}}
\label{eq:P_reflected_camera}
\end{equation}

The number of electrons $N_{\text{e}}$ in the camera sensor generated by the object can be calculated according to \gleichung \ref{eq:N_electrons_camera}. 

\begin{equation}
N_{\text{e}}(R) = \frac{P_{\text{r,camera}}(R)  t_{\text{int}} QE}{h \nu}
\label{eq:N_electrons_camera}
\end{equation}

\gleichung \ref{eq:N_electrons_noise_camera} denotes the calculation of the number of noise electrons $N_{\text{e,n}}$.

\begin{equation}
N_{\text{e,n}} = \sqrt{N_{\text{e}}} + \frac{k B_{\text{n,camera}} T_{\text{sys,camera}} t_{\text{int}} QE } {h \nu}
\label{eq:N_electrons_noise_camera}
\end{equation}

To characterize the sensor performance of the camera, the \snr is used as well, which is determined by the ratio of the generated $N_{\text{e}}$ to the noise electrons $N_{\text{e,n}}$ (\gleichung \ref{eq:SNR_camera}).

\begin{equation}
SNR_{\text{camera}}(R) = \frac{N_{\text{e}}(R)}{N_{\text{e,n}}} 
\label{eq:SNR_camera}
\end{equation}

If the camera is directly exposed to the sun, additional electrons $N_{\text{e,sun}}$ are generated by the incident sunlight, which must be taken into account when calculating the \snr (\gleichung \ref{eq:SNR_camera_sun} and \ref{eq:N_electrons_direct_sun}). 

\begin{equation}
SNR_{\text{camera}}(R) = \frac{N_{\text{e}}(R)}{N_{\text{e,n}} + N_{\text{e,sun}}}
\label{eq:SNR_camera_sun}
\end{equation}


\begin{equation}
N_{\text{e,sun}} = \frac{E_{\text{sun}} A_{\text{r,camera}} t_{\text{int}} QE}{h \nu}
\label{eq:N_electrons_direct_sun}
\end{equation}

\hspace{0.0cm}

\section{METHODOLOGY}
\label{sec:methodology}

\begin{figure}[t]
	\centering
	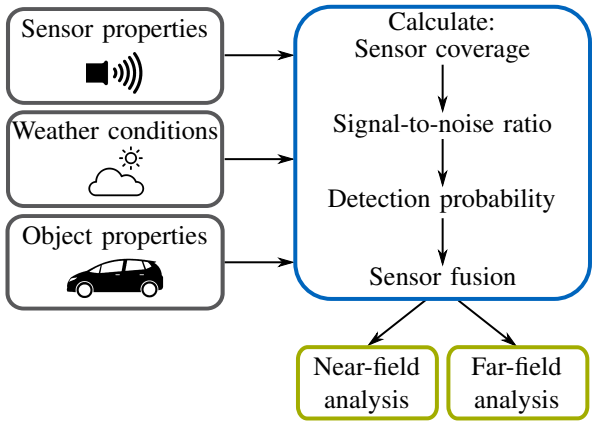
	\caption{Overview of the applied method.}
	\label{fig:methodology}
\end{figure}

For modeling the sensor coverage of automated vehicles, phenomenological models based on \gleichungen \ref{eq:SNR_general} to \ref{eq:N_electrons_direct_sun} are used in this paper. In order to reproduce physical effects phenomenologically, assumptions have to be made for the parameters used in the equations. For sensors currently used in the automotive industry, the necessary information is taken from data sheets. This information as well as weather conditions and properties of the object to be detected serve as input for the applied approach, which is summarized in \abbildung \ref{fig:methodology}. The main part of the method consists of the calculation of the sensor coverage, the \snro, the detection probability and the sensor fusion. These four steps are explained in detail below and they are only calculated within the range of the sensors specified by the manufacturer. Outside this range, the detection probability time is set to zero. As a result, a systematic far and near field analysis can be conducted.

\subsubsection{Sensor Coverage}
\label{subsubsec:sensor_coverage}
Geometric information of the sensors in use is taken for the calculation of the sensor coverage. The theoretical detection range of the sensors is determined on the basis of the field of view and the maximum range. The performance of the sensors is not taken into account. Blind spots in the near field of the vehicle can be analyzed by using the sensor coverage. 

\subsubsection{Signal-to-Noise Ratio}
\label{subsubsec:SNR}
\begin{figure}[!b]
	\centering
	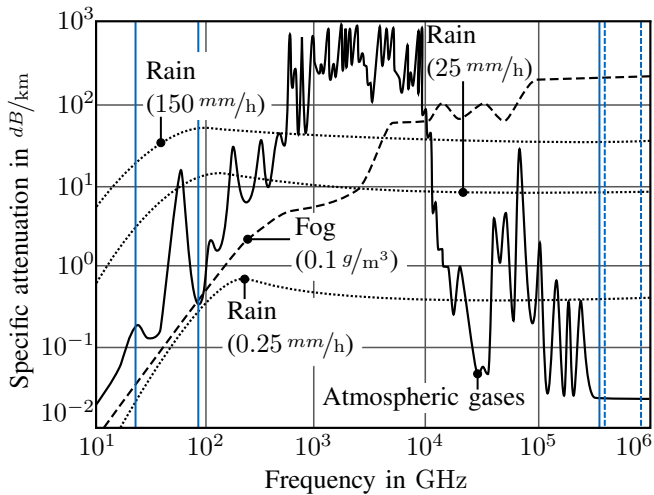
	\caption{Specific attenuation due to atmospheric gases and weather influences based on \cite{Hasirlioglu.2017}. The frequencies of \SI{24}{\giga\hertz} and \SI{77}{\giga\hertz} \radar as well as of \lidar sensors are depicted as vertical solid lines and the dashed lines represent the visible light spectrum.}
	\label{fig:attenuation}
\end{figure}

The next step is to calculate the \snr $SNR_i$ for the used sensors according to \gleichungen \ref{eq:SNR_general} to \ref{eq:N_electrons_direct_sun} from Section \ref{sec:related_work}. Therefore, the specific attenuation due to the atmosphere and weather influences is modeled by constant attenuation factors, which depend on the frequency of the emitted signals. The values of the attenuation factors are based on studies of \cite{Hasirlioglu.2017} (\abbildung \ref{fig:attenuation}). Based on investigations of \cite{Sonbul.2014} an attenuation factor of \SI{1000}{\nicefrac{\decibel}{\kilo\meter}} is assumed for ultrasonic waves. This factor is considered to be weather-independent because of the short range of ultrasonic sensors. 

In Section \ref{sec:related_work}, the equations for the individual active sensors are approximated on the basis of an identical functional principle. Therefore, it is simplified to assume that the cross-section $\sigma$ of the object to be detected is independent of the type of active sensor. In addition, the cross-section $\sigma$ is assumed to be a constant value for simplification, i.e. the orientation of the object has no influence. Based on \textsc{Kamel}'s \cite{Kamel.2017}, \textsc{Matsunami}'s \cite{Matsunami.2012} and \textsc{Schipper}'s \cite{Schipper.2011} investigations for \radar sensors, the cross-sections of the objects are defined as follows: pedestrian $\left(\SI{1}{\square\meter}\right)$, bike $\left(\SI{10}{\square\meter}\right)$, car $\left(\SI{100}{\square\meter}\right)$ and truck $\left(\SI{200}{\square\meter}\right)$. For the passive camera sensor, the geometric cross section is used: pedestrian $\left(\SI{0.9}{\square\meter}\right)$, bike $\left(\SI{1.35}{\square\meter}\right)$, car $\left(\SI{2.7}{\square\meter}\right)$ and truck $\left(\SI{6.75}{\square\meter}\right)$.

\subsubsection{Detection Probability}
\label{subsubsec:detection_probability}

\begin{figure}[!t]
	\centering
	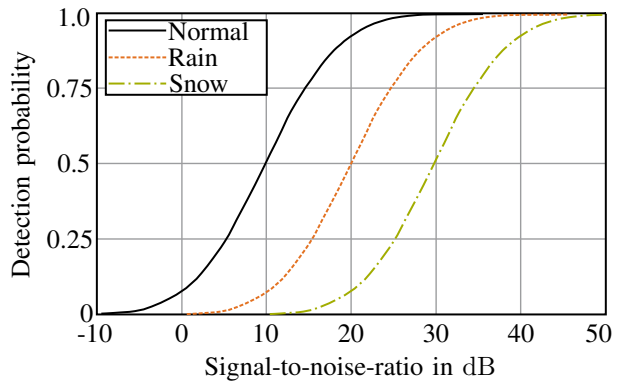
	\caption{Exemplary detection probabilities adapted from \cite{Bernsteiner.2013}. For different weather conditions, modified curves are used to minimize false positives.}
	\label{fig:SNR}
\end{figure}

Based on the calculated \snr values, the detection probability $p_{\text{d},i}$ can be determined. For this purpose, approximations such as the Albersheim equation can be used, which additionally takes into account a false alarm rate, for example. In this paper we use a simplified exemplary conversion of the \snr into a detection probability according to \cite{Bernsteiner.2013} via Receiver Operating Characteristics (ROC) curves (\abbildung \ref{fig:SNR}).

In real applications, the ROC curves are adapted to different environment conditions. In sunny conditions a required \snr value of \SI{10}{\decibel} is considered for a probability of detection of \SI{50}{\percent} \cite{Hasch.2012}.

\subsubsection{Sensor Fusion}
\label{subsubsec:sensor_fusion}
Several areas in the vicinity of the vehicle fall within the detection range of more than one sensor. Therefore, it is necessary to fuse the detection probabilities $p_{\text{d},i}$ of the relevant sensors $i$ in these areas. According to \cite[p. 1465]{Senouci.2012}, this can be achieved by evaluating \gleichung \ref{eq:p_d_fusion}.

\begin{equation}
p_{\text{d,fusion}} = 1 - \prod_{i=1}^{n_{\text{sensors}}} (1-p_{\text{d},i})
\label{eq:p_d_fusion}
\end{equation}

In \gleichung \ref{eq:p_d_fusion}, $p_{\text{d,fusion}}$ denominates the fused detection probability, $n_{\text{sensors}}$ the number of overlapping sensors and $p_{\text{d},i}$ the detection probability of sensor $i$.

As an output of the presented methodology a systematic analysis of the near and far field can be conducted, which is part of the following section.

\section{RESULTS}
\label{sec:results}

This section applies the method explained in Section \ref{sec:methodology} to two different system configurations and compares the results. Both system configurations are based on commercially available vehicles. System A represents a premium vehicle with a balanced sensor setup. System B, on the other hand, increasingly relies on cameras and represents a more cost-effective sensor setup. More information on both system configurations can be found in \tabelle \ref{tab:System_infromation}. 

\begin{table}[!b]
	\vspace{-0.2cm}
	\caption{Summarized information of system configurations A and B. More information can be found in \cite{Mueller.2018}. The horizontal field of view of the sensors is denoted as HFOV.}
	\vspace{-0.2cm}
	\label{tab:System_infromation}
	\begin{center}
		\begin{tabular}{cccrr}
			\toprule 
			& Type & Position 	& HFOV & Range \\
			\midrule
			\multirow{11}{*}{\rotatebox[origin=c]{90}{\hspace{-0.7cm} System A}} & \parbox[t]{10pt}{\multirow{4}{*}{\rotatebox[origin=c]{90}{Camera}}}	& front windshield 		& \ang{50} & \SI{120}{\meter} \\
			&							& front bumper 				& \ang{137} & \SI{60}{\meter} \\
			&							& side mirror left \& right & \ang{137} & \SI{60}{\meter} \\
			&							& rear 						& \ang{137} & \SI{60}{\meter} \\
			\cmidrule[0.5pt]{2-5}
			& \parbox[t]{10pt}{\multirow{3}{*}{\rotatebox[origin=c]{90}{\radarohne}}}	& front bumper center & \ang{30} & \SI{250}{\meter} \\
			&							& front bumper left \& right 	& \ang{150} & \SI{100}{\meter} \\
			&							& rear 	bumper left \& right	& \ang{150} & \SI{100}{\meter} \\
			\cmidrule[0.5pt]{2-5}
			& \parbox[t]{10pt}{\multirow{2}{*}{\rotatebox[origin=c]{90}{Ultras.}}}	& six in front bumper & \ang{70} & \SI{5.5}{\meter} \\ [0.1cm]
			&							& six in rear bumper  	& \ang{70} & \SI{5.5}{\meter} \\
			\cmidrule[0.5pt]{2-5}
			& \parbox[t]{10pt}{\multirow{2}{*}{\rotatebox[origin=c]{90}{\lidarohne}}}	& \multirow{2}{*}{front bumper} & \multirow{2}{*}{\ang{145}} & \multirow{2}{*}{\SI{150}{\meter}} \\ 
			&							&   	&  &  \\ 
			\bottomrule
			\toprule
			\multirow{11}{*}{\rotatebox[origin=c]{90}{\hspace{-0.3cm} System B}} & \parbox[t]{10pt}{\multirow{6}{*}{\rotatebox[origin=c]{90}{Camera}}}	& front windshield 		& \ang{35} & \SI{250}{\meter} \\
			&							& front windshield 					& \ang{45}  & \SI{150}{\meter} \\
			&							& front windshield 					& \ang{120} & \SI{60}{\meter} \\
			&							& forward-facing side cameras		& \ang{90}  & \SI{80}{\meter} \\
			&							& rear-facing side cameras			& \ang{75}  & \SI{100}{\meter} \\
			&							& rear								& \ang{135} & \SI{50}{\meter} \\
			\cmidrule[0.5pt]{2-5}
			& \parbox[t]{10pt}{\multirow{2}{*}{\rotatebox[origin=c]{90}{\radarohne}}}	& \multirow{2}{*}{front bumper} & \multirow{2}{*}{\ang{35}} & \multirow{2}{*}{\SI{160}{\meter}} \\ 
			&							&   	&  &  \\
			\cmidrule[0.5pt]{2-5}
			& \parbox[t]{10pt}{\multirow{2}{*}{\rotatebox[origin=c]{90}{Ultras.}}}	& six in front bumper & \ang{90} & \SI{8}{\meter} \\ [0.1cm]
			&							& six in rear bumper  	& \ang{90} & \SI{8}{\meter} \\
			\bottomrule 
		\end{tabular}
	\end{center}
\end{table}

Both configurations have a maximum sensor range of \SI{250}{\meter} to the front. In System A, this is the \radar sensor in the front bumper, while in System B it is one of the three cameras in the windshield.  To the rear, both systems have a maximum sensor range of approximately \SI{100}{\meter}. Once again, System A relies more on \radar sensors in the rear bumper and System B on rear-facing side cameras. The number of ultrasonic sensors for the detection of near objects is identical in both configurations. Only System A uses a \lidar sensor.

In the following, an analysis of the sensor coverage of the near field around the ego-vehicle is first performed and subsequently one of the far field. When analyzing the far field, different weather conditions are taken into account. 

\subsection{Near-Field Analysis}
\label{subsec:res_NearFieldAnalysis}

\begin{figure*}[t!]
	\centering
	\begin{subfigure}[t]{0.48\textwidth}
		\flushleft
		\includegraphics[scale=1]{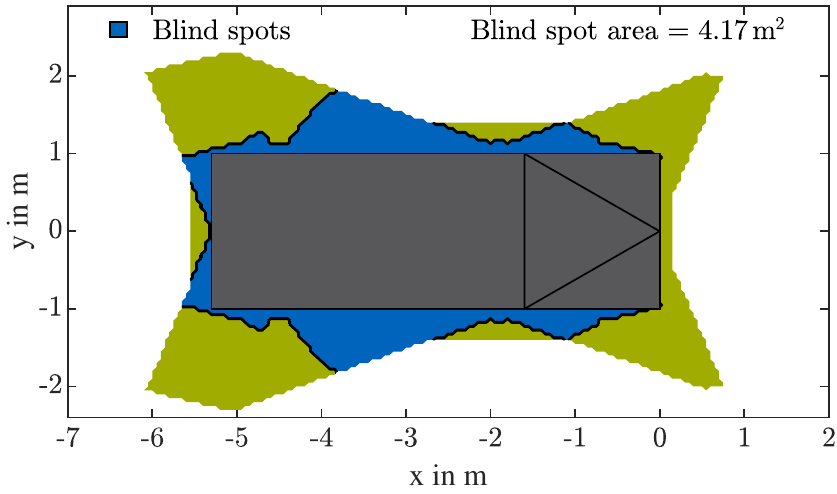}
		\vspace{-0.5cm}
		\caption{System A}
		\label{fig:ANearfield}
	\end{subfigure}%
	~ 
	\begin{subfigure}[t]{0.48\textwidth}
		\flushright
		\includegraphics[scale=1]{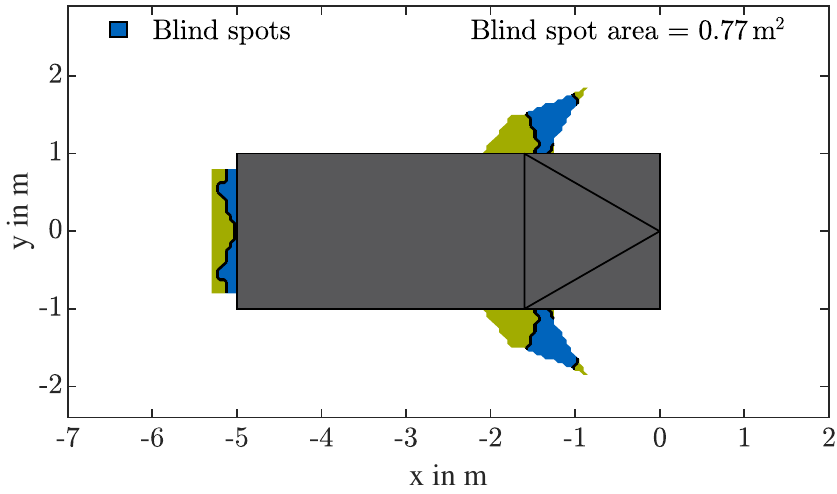}
		\vspace{-0.5cm}
		\caption{System B}
		\label{fig:BNearfield}
	\end{subfigure}
	\vspace{-0.1cm}
	\caption{Blind spots (blue) of both sensor configurations, calculated using data from \tabelle \ref{tab:System_infromation} as well as the vertical opening angles of the sensors from \cite{Mueller.2018}. Blind spots are areas that do not fall within the detection range of a sensor at any height. The sum of all blind spots equals an area of \SI{4.17}{\square\meter} for System A and \SI{0.77}{\square\meter} for System B. In addition, the area that does not fall into any detection area at a height of \SI{0.1}{\meter} is shown in green.}
	\label{fig:Nearfield}
\end{figure*}

For the analysis of the near field, the theoretical detection areas of the individual sensors are superimposed and blind spots are identified. Blind spots are areas that cannot be seen by any sensor at any height. The result for both sensor configurations is depicted in \abbildung \ref{fig:Nearfield}. In addition, the areas that do not fall into any detection area at a height of \SI{0.1}{\meter} are shown in green. A height of \SI{0.1}{\meter} represents the minimal object height to be detected.

%

%

It can be seen that System B has significantly smaller blind spots than System A. This is due to the advantageous distribution of System's B side cameras. By combining front- and rear-facing side cameras, blind spots can be minimized. However, the availability of System's B rear-facing side cameras must be investigated in practice. These are located in the front fender at a height of only \SI{0.9}{\meter}. This makes them particularly susceptible to dust and moisture that can be stirred up while driving. System A, on the other hand, can be more vulnerable to accidents, similar to the accident of an autonomous prototype with a motorcycle \cite{DMV.2017}, due to larger blind spots.

\subsection{Far-Field Analysis}
\label{subsec:res_FarFieldAnalysis}

\begin{figure*}[t!]
	\vspace{-0.4cm}
	\centering
	\begin{subfigure}[t]{0.48\textwidth}
		\flushleft
		\includegraphics[scale=1]{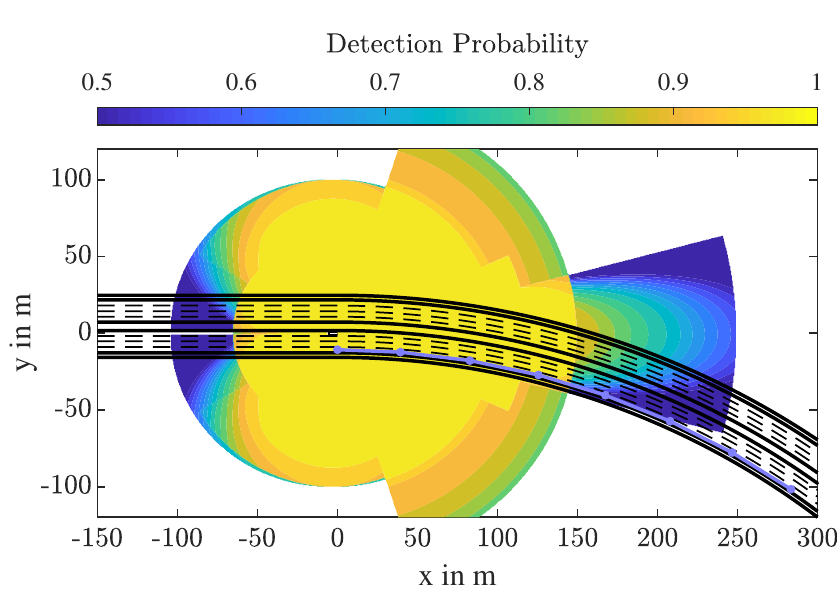}
		\vspace{-0.5cm}
		\caption{System A}
		\label{fig:ASunnyPKW0_75}
	\end{subfigure}%
	~ 
	\begin{subfigure}[t]{0.48\textwidth}
		\flushright
		\includegraphics[scale=1]{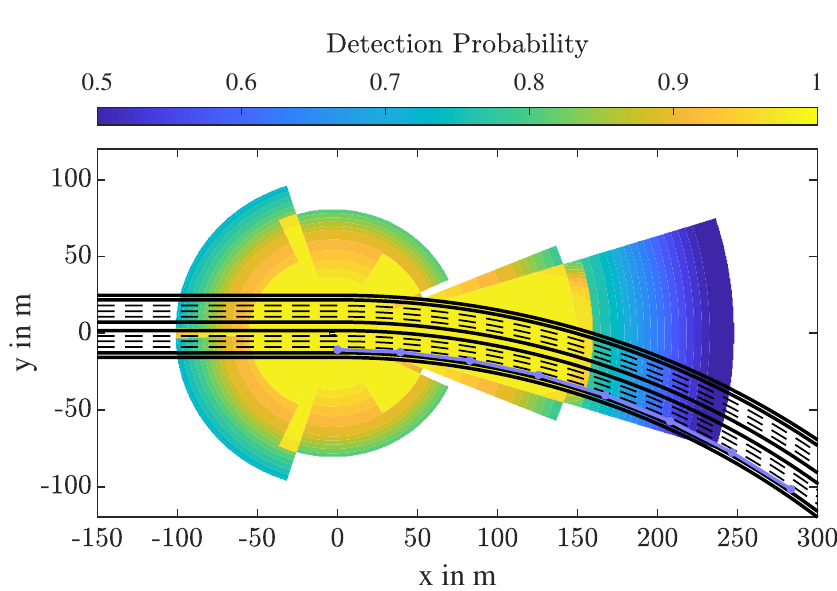}
		\vspace{-0.5cm}
		\caption{System B}
		\label{fig:BSunnyPKW0_75}
	\end{subfigure}
	\vspace{-0.1cm}	
	\caption{Detection probability of the \ego under normal weather conditions, calculated from \gleichung \ref{eq:SNR_general} to \ref{eq:SNR_camera} and \gleichung \ref{eq:p_d_fusion}. The \ego drives in the left lane of a four-lane highway approaching a curve of radius $r=\SI{500}{\meter}$. Depicted in purple is the path of a passenger car driving in the right-hand lane. The path of the passenger car is considered only in the area in front of the automated vehicle.}
	\label{fig:Farfield_sunny}
\end{figure*}

\begin{figure*}[h!]
	\vspace{-0.4cm}
	\centering
	\begin{subfigure}[t]{0.48\textwidth}
		\flushleft
		\includegraphics[scale=1]{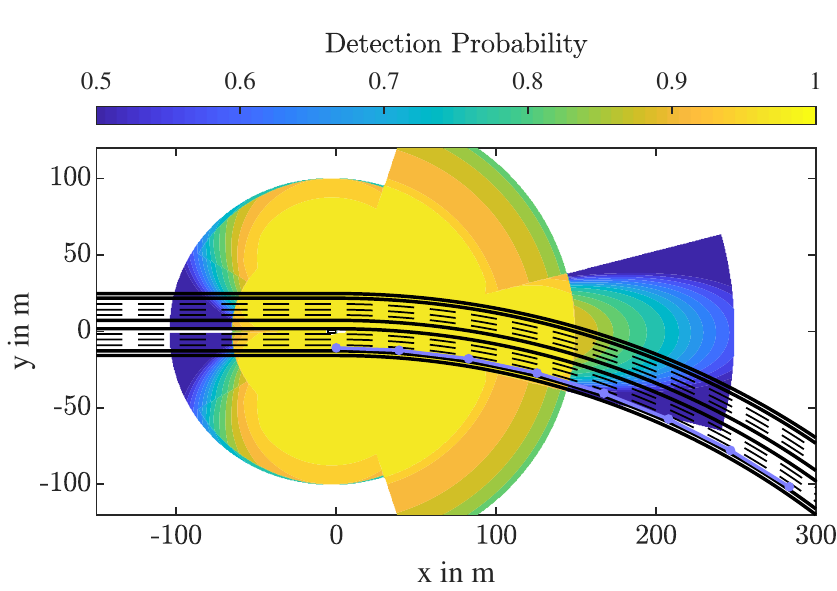}
		\vspace{-0.5cm}
		\caption{System A}
		\label{fig:ABlendungPKW0_75}
	\end{subfigure}%
	~ 
	\begin{subfigure}[t]{0.48\textwidth}
		\flushright
		\includegraphics[scale=1]{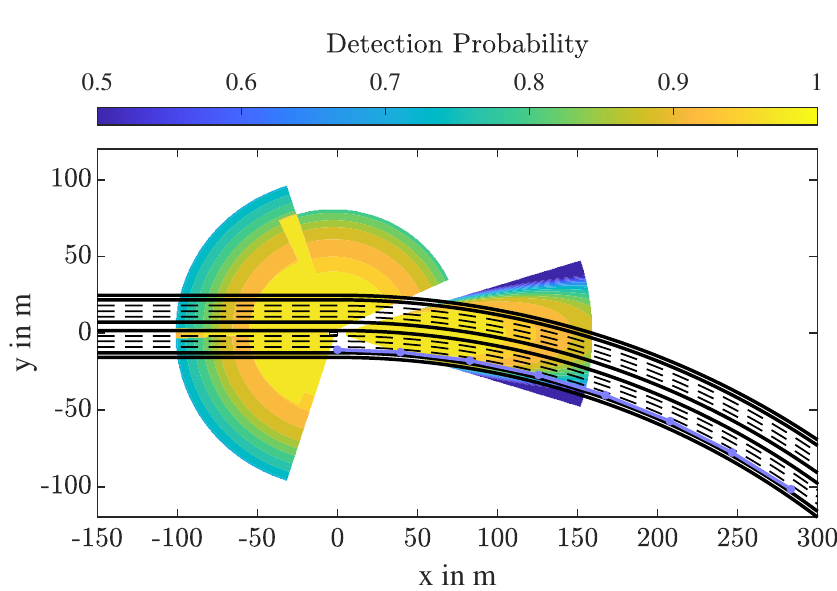}
		\vspace{-0.5cm}
		\caption{System B}
		\label{fig:BBlendungPKW0_75}
	\end{subfigure}
	\vspace{-0.1cm}
	\caption{Detection probability of the \ego with direct glare from the front right, calculated from \gleichung \ref{eq:SNR_general} to \ref{eq:N_electrons_noise_camera} and \gleichung \ref{eq:SNR_camera_sun} to \ref{eq:p_d_fusion}. The \ego drives in the left lane of a four-lane highway approaching a curve of radius $r=\SI{500}{\meter}$. Depicted in purple is the path of a passenger car driving in the right-hand lane. The path of the passenger car is considered only in the area in front of the automated vehicle.}
	\label{fig:Farfield_Blendung}
\end{figure*}

When analyzing the far field, the path of a passenger car through the detection field of the automated vehicle is considered. A four-lane motorway with a curve of radius $r=\SI{500}{\meter}$ is used as an example. For normal ambient conditions, this is shown for both configurations in \abbildung \ref{fig:Farfield_sunny}. As benchmarks, cumulated distances of the passenger car in different detection probability levels in front of the automated vehicle are used (\tabelle \ref{tab:results}). Here and for all subsequent results, the detection probabilities of a plane at a height of $\SI{0.75}{\meter}$ are considered. For comparison, \abbildung \ref{fig:Farfield_Blendung} shows the same scenario with direct sunlight from the front right. The field of view of all forward- and right-facing camera sensors is therefore severely restricted. Again, the results are summarized in \tabelle \ref{tab:results}. The results show that System B performs better than System A under normal environmental conditions. The advantage of using different sensor technologies by System A becomes evident in the direct-sunlight scenario. In this scenario, the performance of System A is considerably better than that of System B. Consequently, the safety assessment of System B must focus more on the consideration of different environmental conditions. 

\begin{table}[b]
	\vspace{-0.2cm}
	\caption{Summary of the results for both system configurations under normal weather conditions and with direct glare from the front right. Shown are the cumulated distances in which the passenger car is within different detection probability $p_{\text{d}}$ levels.}
	\vspace{-0.2cm}
	\label{tab:results}
	\begin{center}
		\begin{tabular}{rrrrr}
			\toprule
			& \multicolumn{2}{c}{Normal}				& \multicolumn{2}{c}{Direct glare} \\
			Detection prob. 			& System A 				& System B 			& System A 				& System B \\
			\midrule
			$p_{\text{d}}=1.0$			& \SI{73}{\meter}  		& \SI{50}{\meter} 	& \SI{73}{\meter}  		& \SI{0}{\meter} 	\\
			$p_{\text{d}}\geq 0.8$ 		& \SI{148}{\meter}  	& \SI{159}{\meter} 	& \SI{148}{\meter}  	& \SI{89}{\meter} 	\\
			$p_{\text{d}}\geq 0.5$ 		& \SI{169}{\meter}  	& \SI{237}{\meter} 	& \SI{169}{\meter}  	& \SI{119}{\meter} 	\\
			$p_{\text{d}}> 0.0$		 	& \SI{210}{\meter}  	& \SI{249}{\meter} 	& \SI{210}{\meter}  	& \SI{121}{\meter} 	\\
			$p_{\text{d}}=0.0$			& \SI{93}{\meter} 		& \SI{54}{\meter} 	& \SI{93}{\meter} 		& \SI{181}{\meter} 	\\		
			\bottomrule
		\end{tabular}
	\end{center}
\end{table}

\section{DISCUSSION}
\label{sec:discussion}

In the presented paper, the phenomenological sensor models taken from the literature are assumed to be validated. This is reasonable, but has to be investigated in detail in future work. Furthermore, detailed information about the sensors used is required (\tabelle \ref{tab:sensor_symbols}). As far as possible, these have been taken from officially accessible data sheets. To improve the achieved results, however, close cooperation with the sensor manufacturers and the OEM is necessary in order to precisely determine all the decisive parameters. 

Further processing of the sensor data to create an environmental map has a considerable influence on the performance of the overall system. This step is not considered in this paper and must be examined in detail in future work.

In addition, only one object is considered in the model used. Therefore, the effects of multiple objects such as shading have not been investigated. An enhancement to represent these effects must be the subject of further work.

\section{CONCLUSION}
\label{sec:conclusion}

This contribution addresses a novel method for the selection of system-specific relevant scenarios for the safety assessment of automated vehicles. This is achieved by using phenomenological sensor models and a subsequent analysis of the position-dependent detection probability of objects. The comparison of two system configurations based on commercially available vehicles shows that a system-specific adaptation of the test scenarios is reasonable. In future work, additional effects such as shading by other objects and further processing of the sensor data must be considered.

\section*{ACKNOWLEDGMENT AND CONTRIBUTIONS}

Thomas Ponn (corresponding author) initiated and wrote this paper. He was involved in all stages of development and primarily developed the research question as well as the concept. Fabian M\"uller wrote his master thesis on sensor modeling and implemented the developed models during his thesis. Frank Diermeyer contributed to the conception of the research project and revised the paper critically for important intellectual content. He gave final approval of the version to be published and agrees to all aspects of the work. As a guarantor, he accepts responsibility for the overall integrity of the paper.

\bibliographystyle{./bibliography/IEEEtran} 
\bibliography{./bibliography/IEEEabrv,./bibliography/mybibfile}

\begin{thebibliography}{10}
\providecommand{\url}[1]{#1}
\csname url@rmstyle\endcsname
\providecommand{\newblock}{\relax}
\providecommand{\bibinfo}[2]{#2}
\providecommand\BIBentrySTDinterwordspacing{\spaceskip=0pt\relax}
\providecommand\BIBentryALTinterwordstretchfactor{4}
\providecommand\BIBentryALTinterwordspacing{\spaceskip=\fontdimen2\font plus
\BIBentryALTinterwordstretchfactor\fontdimen3\font minus
  \fontdimen4\font\relax}
\providecommand\BIBforeignlanguage[2]{{%
\expandafter\ifx\csname l@#1\endcsname\relax
\typeout{** WARNING: IEEEtran.bst: No hyphenation pattern has been}%
\typeout{** loaded for the language `#1'. Using the pattern for}%
\typeout{** the default language instead.}%
\else
\language=\csname l@#1\endcsname
\fi
#2}}

\bibitem{SAEJ3016.2016}
{SAE J3016}, ``Taxonomy and definitions for terms related to on-road motor
  vehicle automated driving systems,'' 2016.

\bibitem{DeutschesZentrumfurLuftundRaumfahrte.V..2018}
\BIBentryALTinterwordspacing
{Deutsches Zentrum f{\"u}r Luft- und Raumfahrt e.\,V.} (2018) Pegasus-project.
  Accessed on: 27.12.2018. [Online]. Available:
  \url{https://www.pegasusprojekt.de/en/home}
\BIBentrySTDinterwordspacing

\bibitem{Althoff.2009}
M.~Althoff, O.~Stursberg, and M.~Buss, ``Safety assessment of driving behavior
  in multi-lane traffic for autonomous vehicles,'' in \emph{2009 IEEE
  Intelligent Vehicles Symposium}, June 2009, pp. 893--900.

\bibitem{Gruber.2018}
F.~Gruber and M.~Althoff, ``Anytime safety verification of autonomous
  vehicles,'' in \emph{2018 21st International Conference on Intelligent
  Transportation Systems (ITSC)}, Nov 2018, pp. 1708--1714.

\bibitem{Madrigal.2017}
\BIBentryALTinterwordspacing
A.~Madrigal. (2017) Inside {W}aymo's secret world for training self-driving
  cars. Accessed on: 27.12.2018. [Online]. Available:
  \url{https://www.theatlantic.com/technology/archive/2017/08/inside-waymos-secret-testing-and-simulation-facilities/537648/}
\BIBentrySTDinterwordspacing

\bibitem{NTSB.2018}
\BIBentryALTinterwordspacing
{National Transportation Safety Board}. (2018) Preliminary report released for
  crash involving pedestrian, {Uber Technologies, Inc}., test vehicle. Accessed
  on: 27.12.2018. [Online]. Available:
  \url{https://www.ntsb.gov/news/press-releases/Pages/NR20180524.aspx}
\BIBentrySTDinterwordspacing

\bibitem{Bernsteiner.2015}
S.~Bernsteiner, Z.~Magosi, D.~Lindvai-Soos, and A.~Eichberger,
  ``{Radarsensormodell f{\"u}r den virtuellen Entwicklungsprozess},''
  \emph{ATZ}, no. 02/2015, pp. 72--79, 2015.

\bibitem{Feilhauer.2017b}
M.~Feilhauer and J.~H{\"a}ring, ``A real-time capable multi-sensor model to
  validate {ADAS} in a virtual environment,'' Wiesbaden, 2017.

\bibitem{Schubert.2014}
R.~Schubert, N.~Mattern, and R.~Bours, ``Simulation of sensor models for the
  evaluation of advanced driver assistance systems,'' \emph{ATZ}, no. 03/2014,
  pp. 26--29, 2014.

\bibitem{Cao.2017}
P.~Cao, ``Modeling active perception sensors for real-time virtual validation
  of automated driving systems,'' Dissertation, {Technische Universit{\"a}t
  Darmstadt}, Darmstadt, 2017.

\bibitem{Steinbaeck.2017}
J.~Steinbaeck, C.~Steger, G.~Holweg, and N.~Druml, ``Next generation radar
  sensors in automotive sensor fusion systems,'' in \emph{2017 Sensor Data
  Fusion: Trends, Solutions, Applications (SDF)}, Oct 2017, pp. 1--6.

\bibitem{Schrepfer.2018}
J.~Schrepfer, J.~Mathes, V.~Picron, and H.~Barth, ``{Automatisiertes Fahren und
  seine Sensorik im Test},'' \emph{ATZ}, no. 01/2018, pp. 28--37, 2018.

\bibitem{Sonbul.2014}
O.~Sonbul and A.~Kalashnikov, ``Determining the operating distance of air
  ultrasound range finders: Calculations and experiments,'' \emph{International
  Journal of Computing}, no. 13 (2), pp. 125--131, 2014.

\bibitem{Grgic.2015}
M.~Grgic, ``Generic radar model for automotive applications,'' Master's thesis,
  {TU Graz}, Graz, 2015.

\bibitem{Ludloff.1998}
A.~Ludloff, \emph{Praxiswissen Radar und Radarsignalverarbeitung}.\hskip 1em
  plus 0.5em minus 0.4em\relax {Vieweg + Teubner}, 1998.

\bibitem{Blake.1986}
L.~V. Blake, \emph{Radar Range-Performance Analysis}.\hskip 1em plus 0.5em
  minus 0.4em\relax {Artech House}, 1986.

\bibitem{Richmond.2010}
R.~D. Richmond and S.~C. Cain, \emph{Direct-detection LADAR systems}.\hskip 1em
  plus 0.5em minus 0.4em\relax {SPIE Press}, 2010.

\bibitem{TristanK..2010}
C.~{Tristan K.}, S.~{K. Clint}, C.~{William E.}, and S.~{Kris Y.}, ``Predicting
  small target detection performance of low-snr airborne lidar,'' \emph{IEEE
  Journal of Selected Topics in Applied Earth Observations and Remote Sensing},
  no. Vol. 3, No. 4, pp. 672--688, 2010.

\bibitem{Sameh.2016}
A.~Sameh, S.~David, A.~Mark, M.~Fred, and A.~Sam, ``Signal to noise ratio
  characterization of coherrent doppler lidar backscattered siganls.''

\bibitem{Hayat.2000}
M.~Hayat and G.~Dong, ``A new approach for computing the bandwidth statistics
  of avalanche photodiodes,'' \emph{IEEE Transactions on Electron Devices}, no.
  Vol. 47, No. 6, pp. 1273--1279, 2000.

\bibitem{Rudolph.2017}
\BIBentryALTinterwordspacing
G.~Rudolph and U.~Voelzke. (2017) Three sensor types drive autonomous vehicles.
  Accessed on: 27.12.2018. [Online]. Available:
  \url{https://www.sensorsmag.com/components/three-sensor-types-drive-autonomous-vehicles}
\BIBentrySTDinterwordspacing

\bibitem{Lange.2000}
R.~Lange, ``3d time-of-flight distance measurement with custom solid-state
  image sensors in cmos/ccd-technology,'' Dissertation, {Universit{\"a}t
  Siegen}, Siegen, 2000.

\bibitem{Hasirlioglu.2017}
S.~Hasirlioglu and A.~Riener, ``Introduction to rain and fog attenuation on
  automotive surround sensors,'' in \emph{2017 IEEE 20th International
  Conference on Intelligent Transportation Systems (ITSC)}, Oct 2017.

\bibitem{Kamel.2017}
E.~Kamel, A.~Peden, and P.~Pajusco, ``{RCS Modeling and Measurements for
  Automotive Radar Applications in the W Band},'' \emph{11th European
  Conference on Antennas and Propagation}, pp. 2445--2449, 2017.

\bibitem{Matsunami.2012}
I.~Matsunami, R.~Nakamura, and A.~Kajiwara, ``{RCS measurements for vehicles
  and pedestrian at 26 and 79 GHz},'' in \emph{2012 6th International
  Conference on Signal Processing and Communication Systems}, Dec 2012, pp.
  1--4.

\bibitem{Schipper.2011}
T.~Schipper, J.~Fortuny-Guasch, D.~Tarchi, L.~Reichardt, and T.~Zwick, ``{RCS
  Measurement Results for Automotive Related Objects at 23-27 GHz},''
  \emph{Proceedings of the 5th European Conference on Antennas and
  Propagation}, pp. 683--686, 2011.

\bibitem{Bernsteiner.2013}
S.~Bernsteiner, Z.~Magosi, D.~Lindvai-Soos, and A.~Eichberger,
  ``\BIBforeignlanguage{deutsch}{Ph{\"a}nomenologisches {R}adarsensormodell zur
  {S}imulation l{\"a}ngsdynamisch regelnder {F}ahrerassistenzsysteme},'' in
  \emph{\BIBforeignlanguage{deutsch}{16. Internationaler Kongress Elektronik im
  Fahrzeug}}, ser. VDI-Berichte, vol. 2188.\hskip 1em plus 0.5em minus
  0.4em\relax Springer-VDI-Verlag GmbH \& Co.KG, 2013, pp. 639--650.

\bibitem{Hasch.2012}
J.~Hasch, E.~Topak, R.~Schnabel, T.~Zwick, R.~Weigel, and C.~Waldschmidt,
  ``Millimeter-wave technology for automotive radar sensors in the 77 {GHz}
  frequency band,'' \emph{IEEE Transactions on Microwave Theory and
  Techniques}, no. Vol. 60, No. 3, pp. 845--860, 2012.

\bibitem{Senouci.2012}
M.~Senouci, A.~Mellouk, L.~Oukhellou, and A.~Aissani, ``An evidence-based
  sensor coverage model,'' \emph{IEEE Communications Letters}, no. Vol. 16, No.
  9, pp. 1462--1465, 2012.

\bibitem{Mueller.2018}
F.~M{\"u}ller, ``{Modellierung der Sensorabdeckung autonomer Fahrzeuge zur
  Berechnung optimaler Ann{\"a}herungspfade},'' Master's thesis, {Technische
  Universit{\"a}t M{\"u}nchen}, M{\"u}nchen, 2018.

\bibitem{DMV.2017}
\BIBentryALTinterwordspacing
{Department of Motor Vehicles}. (2017) Report of traffic accident involving an
  autonomous vehicle. Accessed on: 27.12.2018. [Online]. Available:
  \url{\url{https://www.dmv.ca.gov/portal/wcm/connect/1877d019-d5f0-4c46-b472-78cfe289787d/GMCruise{\_}120717.pdf?MOD=AJPERES}}
\BIBentrySTDinterwordspacing

\end{thebibliography}

\end{document}